\documentclass[10pt,conference]{IEEEtran}

\usepackage{cite}

\ifCLASSINFOpdf
   \usepackage[pdftex]{graphicx}
   \graphicspath{{figs/}}
   \DeclareGraphicsExtensions{.pdf,.jpeg,.png}
\else
   \usepackage[dvips]{graphicx}
   \graphicspath{{../figs/}}
   \DeclareGraphicsExtensions{.eps}
\fi

\usepackage[cmex10]{amsmath}
\usepackage{amsthm}

\usepackage{algorithmic}

\usepackage{array}

\ifCLASSOPTIONcompsoc
\usepackage[caption=false,font=normalsize,labelfont=sf,textfont=sf]{subfig}
\else
  \usepackage[caption=false,font=footnotesize]{subfig}
\fi

\usepackage{url}

\usepackage{comment}
\usepackage{booktabs}
\usepackage{multirow}
\usepackage{xcolor}
\usepackage[normalem]{ulem}
\usepackage{soul}

\def\baselinestretch{0.98}

\newcommand{\ie}{\textit{i.e.}}

\newcommand{\etal}{\textit{et~al.}}
\newcommand*{\ra}{\ensuremath{\rightarrow}}

\newif\ifreview
\reviewtrue

\ifreview
\newcommand{\rv}[2][]{%
    \if\relax\detokenize{#1}\relax%
        \textcolor{blue}{#2}%
    \else%
        \textcolor{gray}{\sout{#1}}\textcolor{red}{#2}%
    \fi%
}
\newcommand{\todo}[1]{\noindent\textcolor{orange}{To-Do: {#1}}}
\else
\newcommand{\rv}[2][]{#2}
\newcommand{\todo}[1]{\empty}
\fi

\newif\iffinal
\finaltrue
\newcommand{\cmtid}{100}

\iffinal
\else
\usepackage[switch]{lineno}
\fi

\begin{document}
%
\thispagestyle{empty}
\onecolumn
\linespread{1.2}\selectfont{}
{\noindent\Huge IEEE Copyright Notice}\\[1pt]

{\noindent\large Copyright (c) 2023 IEEE

\noindent Personal use of this material is permitted. Permission from IEEE must be obtained for all other uses, in any current or future media, including reprinting/republishing this material for advertising or promotional purposes, creating new collective works, for resale or redistribution to servers or lists, or reuse of any copyrighted component of this work in other works.}\\[1em]

{\noindent\Large Accepted to be published in: 2023 36th SIBGRAPI Conference on Graphics, Patterns and Images (SIBGRAPI'23), November 6--9, 2023.}\\[1in]

{\noindent\large Cite as:}\\[1pt]

{\setlength{\fboxrule}{1pt}
 \fbox{\parbox{0.65\textwidth}{L. F. A. Silva, N. Sebe, and J. Almeida, ``Tightening Classification Boundaries in Open Set Domain Adaptation through Unknown Exploitation'' in \emph{2023 36th SIBGRAPI Conference on Graphics, Patterns and Images (SIBGRAPI)}, Rio Grande, RS, Brazil, 2023, pp. 1--6}}}\\[1in] 
 
{\noindent\large BibTeX:}\\[1pt]

{\setlength{\fboxrule}{1pt}
 \fbox{\parbox{0.95\textwidth}{
 @InProceedings\{SIBGRAPI\_2023\_Silva,
 
 \begin{tabular}{lll}
  & author    & = \{L. F. A. \{Silva\} and 
                    N. \{Sebe\} and
                    J. \{Almeida\}\},\\
			   
  & title     & = \{Tightening Classification Boundaries in Open Set Domain Adaptation \\
  &           & \ \ through Unknown Exploitation\},\\
			   
  & pages     & = \{1--6\},\\
  
  & booktitle & = \{2023 36th {SIBGRAPI} Conference on Graphics, Patterns and Images ({SIBGRAPI})\},\\
  
  & address   & = \{Rio Grande, RS, Brazil\},\\
  
  & month     & = \{November 6--9\},\\
  
  & year      & = \{2023\},\\
  
  & publisher & = \{\{IEEE\}\},\\
  
  \end{tabular}
  
\}
 }}}

\twocolumn
\linespread{1}\selectfont{}
\clearpage

\title{Tightening Classification Boundaries in Open Set Domain Adaptation through Unknown Exploitation}

\iffinal

\author{
\IEEEauthorblockN{
Lucas Fernando Alvarenga e Silva$^1$,
Nicu Sebe$^2$,
Jurandy Almeida$^3$
}\\
\IEEEauthorblockA{
$^1$\textit{IC/University of Campinas (UNICAMP) -- SP, Brazil}\\
$^2$\textit{DISI/University of Trento (UniTN) -- TN, Italy}\\
$^3$\textit{DC/Federal University of S\~{a}o Carlos (UFSCar) -- SP, Brazil}\\
Emails: {\small\texttt{lucas.silva@ic.unicamp.br, niculae.sebe@.unitn.it, jurandy.almeida@ufscar.br}}
}
}
\else
  \author{Sibgrapi paper ID: \cmtid \\ }
  \linenumbers
\fi

\maketitle

\begin{abstract}
Convolutional Neural Networks~(CNNs) have brought revolutionary advances to many research areas due to their capacity of learning from raw data.
However, when those methods are applied to non-controllable environments, many different factors can degrade the model’s expected performance, such as unlabeled datasets with different levels of domain shift and category shift.
Particularly, when both issues occur at the same time, we tackle this challenging setup as Open Set Domain Adaptation~(OSDA) problem.
In general, existing OSDA approaches focus their efforts only on aligning known classes or, if they already extract possible negative instances, use them as a new category learned with supervision during the course of training.
We propose a novel way to improve OSDA approaches by extracting a high-confidence set of unknown instances and using it as a hard constraint to tighten the classification boundaries of OSDA methods.
Especially, we adopt a new loss constraint evaluated in three different means, (1) directly with the \textit{pristine} negative instances; (2) with randomly \textit{transformed} negatives using data augmentation techniques; and (3) with synthetically \textit{generated} negatives containing adversarial features.
We assessed all approaches in an extensive set of experiments based on OVANet, where we could observe consistent improvements for two public benchmarks, the Office-31 and Office-Home datasets, yielding absolute gains of up to 1.3\% for both Accuracy and H-Score on Office-31 and 5.8\% for Accuracy and 4.7\% for H-Score on Office-Home.
\end{abstract}

\IEEEpeerreviewmaketitle

\section{Introduction}

In the last few years, Deep Learning~(DL) methods have brought revolutionary results to several research areas, mainly because of the Convolutional Neural Networks~(CNNs) in computer-vision-related problems~\cite{TPAMI_2020_Geng}.
In general, those methods are expected to work under unrealistic assumptions, such as a controlled environment with a totally labeled set of data and a Closed Set~(CS) of categories~\cite{TPAMI_2020_Geng}.
However, this assumption does not usually hold in uncontrollable environments, like in real-world problems.
Indeed, the process of labeling data is expensive, time-consuming, or in some cases even impossible, and DL models that are supposed to work under supervised datasets generally struggle when faced with unlabeled or partially labeled datasets~\cite{ECCV_2020_Bucci, CVIU_2022_Saltori}.

\begin{figure}[!htb]
    \centering
    \includegraphics[width=0.45\textwidth]{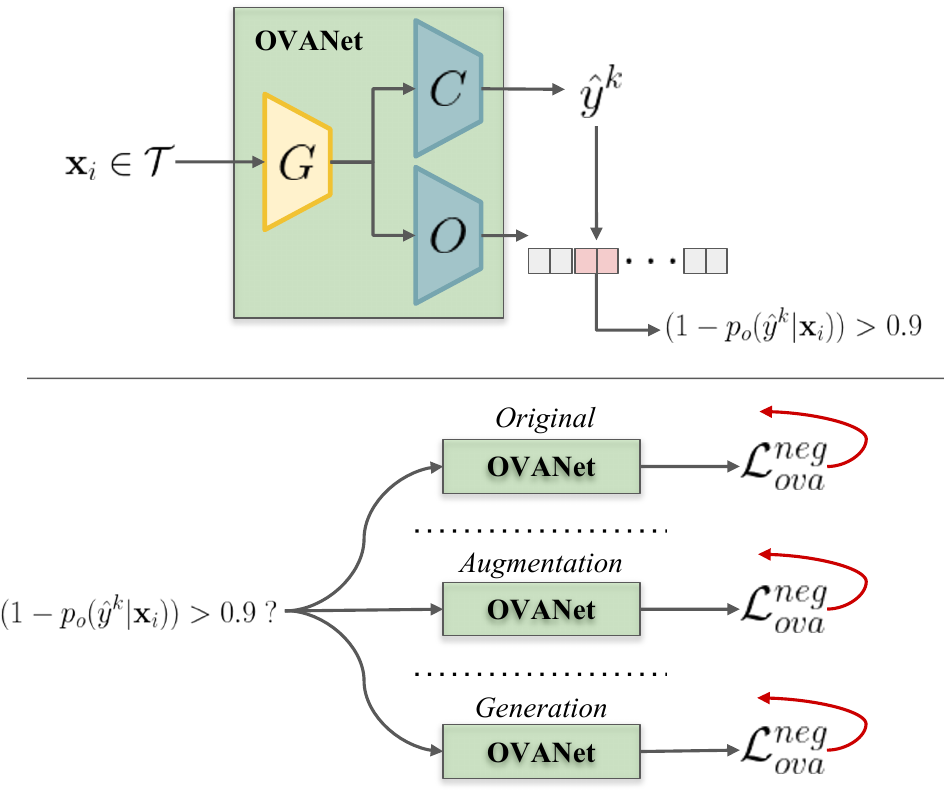}
    \caption{High-level picture of our proposal. In the upper part of the figure, we show the unknown instance extraction step from the pre-trained OVANet, which is used to tighten the boundaries of OVANet by a new constraint $\mathcal{L}_{ova}^{neg}$ added to the loss function. Following, in the bellow part, we evaluate the use of negatives in a three-way manner, namely: Original, Augmentation, and Generation.}
    \label{fig:overview}
\end{figure}

In uncontrollable environments, we can be faced with two main problems: the decreased level of supervision, and no control about the incoming data.
For the first problem, it is usual to use a completely annotated dataset (\ie~source domain), which should be similar to the unlabeled target dataset (\ie~target domain), to train the model. 
However, since the data may come from different underlying distributions, it can induce the domain-shift problem~\cite{SIBGRAPI_2021_Silva}.
For the second problem, our training and test data can present some level of category-shift since we do not know the data beforehand, requiring the model to handle examples from possible unknown/unseen categories during inference~\cite{TPAMI_2021_Chen}.
Each of these problems has its own research areas, the Unsupervised Domain Adaptation~(UDA) and Open Set~(OS) recognition, that individually aim to tackle and reduce domain-shift and category-shift problems, respectively.
Recently, Busto~\etal~\cite{ICCV_2017_Busto} were the pioneers to describe the Open Set Domain Adaptation~(OSDA), the more realistic and challenging scenario where both problems occur at the same time.
The OSDA has been positively accepted as a new research field in the literature with many different contributions, for instance, with approaches based on adversarial techniques~\cite{ECCV_2018_Saito, ECCV_2020_Rakshit, NC_2020_Gao, WACV_2022_Baktashmotlagh}, Extreme Value Theory (EVT) modeling techniques~\cite{bigdata_2021_Xu}, self-supervised tecniques~\cite{ECCV_2020_Bucci}, contrastive learning approaches~\cite{Bucci_2022_WACV}, and gradient-based analysis~\cite{ECCV_2022_Liu}.

Most OSDA methods rely on rejecting unknown samples in target domain and aligning known samples in the target domain with the source domain. For this, they draw known and unknown sets of the high-confidence samples from the target domain. To alleviate the domain shift, positive samples from the known set are aligned with the source domain~\cite{ECCV_2020_Bucci, ECCV_2020_Rakshit, Bucci_2022_WACV}.
On the other hand, negative samples from the unknown set are usually underexploited during the training, being often assigned to an additional logit of the classifier representing the unknown category, which is learned with supervision along with known classes~\cite{ECCV_2022_Liu}.
Liu~\etal~\cite{ECCV_2022_Liu} noted that the unknown set of the target domain is highly informative, possessing complicated semantics and possible correlation to known classes that can hind the oversimplified approaches.
Thus, some recent works~\cite{WACV_2022_Baktashmotlagh, ECCV_2022_Liu} have leveraged such challenging information to improve their OSDA approaches.

Following recent work based on unknown exploitation~\cite{ECCV_2022_Liu, WACV_2022_Baktashmotlagh} and closed-set relationship with OS~\cite{ICLR_2022_Vaze}, we hypothesize that using the unknown set of the target domain to tighten the boundaries of the closed-set classifier can further improve the classification performance.
Thus, we investigate this assumption using the OVANet~\cite{Saito_2021_ICCV} approach, a UNiversal Domain Adaptation~(UNDA) method, that we extended in a three-way manner in our evaluations for the OSDA setting.
We first extract high-confidence negatives from the target domain based on a higher confidence threshold~\cite{ECCV_2020_Rakshit} in order to (1) evaluate the direct use of pure negatives as a new constraint for the known classification (\textbf{original} approach); (2) use data augmentation to randomly transform negatives before applying them as the classification constraint (\textbf{augmentation} approach); and (3) create negative/adversarial examples by a Generative Adversarial Network~(GAN) model trained with such negatives, whose objective is to deceive the OVANet by posing synthetic instances as positive samples that it must later learn to reject (\textbf{generation} approach).
We conducted an extensive set of experiments for the object recognition problem with the Office-31~\cite{ECCV_2010_Saenko} and Office-Home~\cite{CVPR_2017_Venkateswara} datasets.
Our results indicate that such approaches may increase performance in most OSDA tasks, reaching up to 1.3\% of absolute gains for both Accuracy and H-Score on Office-31 and 5.8\% for Accuracy and 4.7\% for H-Score on Office-Home.

The remainder of this paper is organized as follows. Section~\ref{sec:related-work} presents related work. Section~\ref{sec:negatives} describes our approach to deal with negative samples. Section~\ref{sec:experiments} presents the experimental setup and results in two challenging datasets. Finally, in Section~\ref{sec:conclusion} we present the conclusions and directions for future research.

\section{Related Work} \label{sec:related-work}



OSDA mixes both OS and UDA, a more realistic scenario where irrelevant categories in the target domain do not appear in source domain and distributions of partially shared categories between source and target domains are not well aligned.
Despite the vast literature on OS~\cite{Neal_2018_ECCV,TPAMI_2021_Chen,ICLR_2022_Vaze} and UDA~\cite{WVC_2020_Silva,Saito_2020_NeurIPS,Saito_2021_ICCV,SIBGRAPI_2021_Silva}, OSDA has so far been little-studied.

OSDA was introduced by the pioneering work of Busto~and~Gall~\cite{ICCV_2017_Busto} and further defined and described by Saito~\etal~\cite{ECCV_2018_Saito}.
Most OSDA methods~\cite{ECCV_2020_Bucci, ICCV_2017_Busto, ECCV_2018_Saito, bigdata_2021_Xu} only deal with one source domain, but it is possible to extend them to the multi-source setting~\cite{ECCV_2020_Rakshit, Bucci_2022_WACV}.
Bucci~\etal~\cite{ECCV_2020_Bucci} proposed a self-supervised model named ROS that leverages relative-rotation and multi-rotation proxy tasks to close the domain gap and learn relevant features from both domains.
Later, Bucci~\etal~\cite{Bucci_2022_WACV} took one step further and proposed HyMOS, an approach to tackle the multi-source OSDA, which combines contrastive learning and hyperspherical learning to help the source-to-source alignment and style transfer to help source-to-target alignment.
Rakshit~\etal~\cite{ECCV_2020_Rakshit} proposed MOSDANET, an adversarial method that tackles the multi-source setting of OSDA. 
Recently, Liu~\etal~\cite{ECCV_2022_Liu} and Baktashmotlagh~\etal~\cite{WACV_2022_Baktashmotlagh} observed that unknown examples are complex structures that are generally overlooked as a simple new logit added to the classifier.
Motivated by this insight, Liu~\etal\cite{ECCV_2022_Liu} proposed UOL, an approach that leverages a multi-unknown detector equipped with weight discrepancy constraint and gradient-graph induced annotation to design a feature space that can learn the complex semantics of the unknown samples, while Baktashmotlagh~\etal~\cite{WACV_2022_Baktashmotlagh} proposed to generate source-like negative instances by a GAN and use it as a source closed/target supervision.


In general, previous OSDA methods exploit thresholding strategies to reject unknown examples. MOSDANET, ROS, and HyMOS draw possibly known instances from the target domain to align their distribution with source domain during training and unknown examples as supervision for the extra logit of the classifier linked to the unknown category.
On the other hand, we propose to only deal with the unknown subset, extracting the most high-confidence instances from the target domain and using them to tighten known classification boundaries from the OVANet approach.

\section{Dealing with Negatives} \label{sec:negatives}

This section presents our approaches. At first, we describe the confidence measure and the extraction procedure adopted for drawing a set of likely unknown examples from the target domain. In sequence, our strategies based on original unknown instance usage, data augmentation, and instance generation are depicted, showing how we exploit valuable information from negative samples and improve the class separability of the OVANet method.

To assess our unknown-oriented learning strategies, we adopted OVANet~\cite{Saito_2021_ICCV}, an UNDA approach used as the basis of all investigations carried out in this work.
OVANet is a recent work that achieved good results on all four possible class alignments of UNDA: closed-set, partially aligned for closed-set, partially aligned for open-set, and completely open-set fronts. 
The authors are well recognized for their excellence in this field, also being the authors of past relevant works such as OSDA+BP~\cite{ECCV_2018_Saito}, one of the first works that adopted a stronger and more challenging definition for OSDA, and DANCE~\cite{Saito_2020_NeurIPS}, the work that defined the UNDA generalization. 
Moreover, its source code is freely available on the internet\footnote{\url{https://github.com/VisionLearningGroup/OVANet}}.

In the OSDA problem, we are given as input a labeled source domain $\mathcal{D}_s$ sharing a set $L_s$ of categories with an unlabeled target domain $\mathcal{D}_t$ which also includes an additional set $L_{unk}$ of categories, such that $L_t = L_s \cup L_{unk}$. 
The source domain $\mathcal{D}_s = \{ ( \mathbf{x}_i^{s}, y_i^{s}) \}_{i=1}^{N_s}$ is a set of tuples composed of $N_s$ samples $\mathbf{x}_i^{s}$ and their respective labels $y_i^{s}$. 
Since the target domain labels are not known beforehand, the set $\mathcal{D}_t = \{ \mathbf{x}_i^{t} \}_{i=1}^{N_t}$ contains only $N_t$ target samples.
Thus, a sample $\mathbf{x}_i^{t}$ is said to be ``known'' if it relates to some category in $L_s$, otherwise it is said to be ``unknown'' and relates to $L_{unk}$.
UNDA and OSDA methods aim to correctly classify known samples into categories of $L_s$ while rejecting possibly unknown samples that would be assigned to some ``unknown'' category of $L_{unk}$.

OVANet is a method that exploits inter-class distance to learn tight class boundaries for the known samples.
It uses a feature extractor $G$ shared between a closed-set classifier $C$ and a set $O$ of open-set binary classifiers (Figure~\ref{fig:overview}).
Roughly speaking, $C$ is a linear classifier trained using cross-entropy loss in order to correctly classify samples into categories of $L_s$.
On the other hand, $O$ is a set of $|L_s|$ one-vs-all binary classifiers, one for each category in $L_s$, \ie, $O = \{ B_i \}_{i=1}^{|L_s|}$, where the binary classifier $B_i$ is trained to recognize instances from its related category as ``known'' and instances for other categories as ``unknown''. It is achieved through the use of \textit{Hard Negative Classifier Sampling} and \textit{Open-set Entropy Minimization} loss functions (see Equation~2 of \cite{Saito_2021_ICCV}).
Finally, we denote $p_c(y^k|\mathbf{x}_i)$ as the probability that the classifier $C$ associates a sample $\mathbf{x}_i$ to the $k$-th category, and $p_o(\hat{y}^k|\mathbf{x}_i)$ as the probability that the same instance is ``known'' for the binary classifier $B_k$ of $O$ and, therefore, it is ``unknown'' for $B_k$ with probability $1-p_o(\hat{y}^k|\mathbf{x}_i)$.

During inference, the incoming instance $\mathbf{x}_i^{t}$ is assigned to a pseudo-label $\hat{y}_i^{t}$ based on the maximum probability of the classifier $C$, \ie, $\hat{y}_i^{t} = \mathrm{argmax}_{k}( p_c(y^k|\mathbf{x}_i^{t}) )$.
Then, we assert whether the pseudo-label $\hat{y}_i^{t}$ is ``known'' or ``unknown'' based on the open-set classifier.
If $p_o(\hat{y}_i^{t}|\mathbf{x}_i^{t}) \geq 0.5$ then $\hat{y}_i^{t}$ is considered the correct label, otherwise the instance is treated as ``unknown'' and rejected.

Given the inference procedure of OVANet, we hypothesize that leveraging knowledge from negative samples can strengthen the model's learning ability and, consequently, tighten the known/unknown classification boundaries.
To achieve this, we propose to train OVANet for half of the expected epochs, extract the most confident unknown examples from the target domain using the actual OVANet knowledge, and, finally, use them as an additional negative supervision to fine-tune the binary classifiers in $O$ for the remaining epochs.
Particularly, we extract a subset $\mathbf{\overline{X}} \subset \mathcal{D}_t$ of likely negative samples by adopting a very-high threshold to treat a sample as ``unknown'', \ie, $\mathbf{\overline{X}} = \{ \mathbf{x}_i^{t} | \mathbf{x}_i^{t} \in \mathcal{D}_t, 1-p_o(\hat{y}_i^{t} | \mathbf{x}_i^{t}) > 0.9 \}$.
Then, we use this subset as a source of information to force the binary classifiers in $O$ to tight the known/unknown classification boundaries by adding the following constraint in the loss function of OVANet (see Equation~2 of~\cite{Saito_2021_ICCV}):
\begin{equation} \label{eq:loss_ova_neg}
    \mathcal{L}_{ova}^{neg} = -\frac{1}{|L_s|}\sum_{k=1}^{|L_s|}\log(1-p_o(\hat{y}^k|\mathbf{\overline{x}}_i))
\end{equation}

\begin{figure}[!t]
    \centering
\centering
\subfloat[Original]{\includegraphics[width=0.15\textwidth]{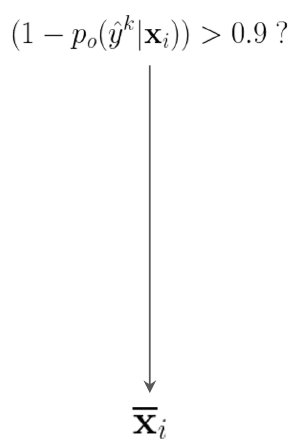}%
\label{fig:approaches:original}}
\hfil
\subfloat[Augmentation]{\includegraphics[width=0.15\textwidth]{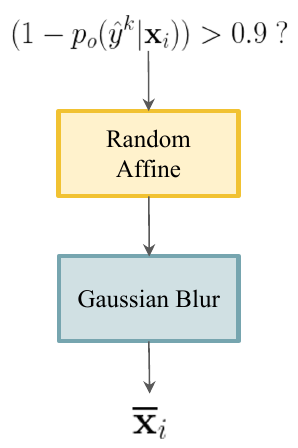}%
\label{fig:approaches:augmentation}}
\hfil
\subfloat[Generation]{\includegraphics[width=0.15\textwidth]{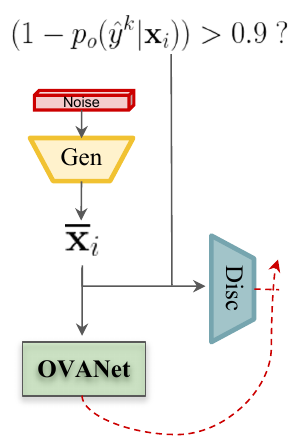}%
\label{fig:approaches:generation}}
\caption{Description of all three approaches. (a) shows the \textit{original} setting, where unknown examples are directly used during OVAnet last training steps. (b) shows the \textit{augmentation} setting, which applies two transformations before the continuation of OVANet training. (c) shows the generation approach, in which a DCGAN is trained to mimic the unknown instances and fool the actual OVANet. The red arrow in (c) shows that Gen module is optimized by errors calculated from both Disc and OVANet structures.}
\label{fig:approaches}
\end{figure}

We assessed 3 different ways to leverage such information:
\begin{enumerate}
    \item \textbf{Original} (Figure~\ref{fig:approaches:original}): Use the negative samples in $\mathbf{\overline{X}}$ to adjust the binary classifiers in $O$.
    \item \textbf{Augmentation} (Figure~\ref{fig:approaches:augmentation}): Apply data augmentation techniques to randomly transform the negative samples in $\mathbf{\overline{X}}$ before feeding them to the binary classifiers of $O$. In general, this approach keeps the same main features as the pristine instances, except by small perturbations added by the transformations.
    \item \textbf{Generation} (Figure~\ref{fig:approaches:generation}): Use a GAN to generate synthetic samples similar to negative samples of $\mathbf{\overline{X}}$. For this, we propose to interrupt the training of OVANet after half of the epochs, then train a GAN for a certain number of epochs using $\mathbf{\overline{X}}$, and, finally, resume the OVANet training but using the synthetically generated samples to adjust the binary classifiers in $O$. This strategy allows us to have more control over features of interest. Roughly speaking, the GAN is composed of a discriminator and a generator that are jointly and adversarially trained. The discriminator is often trained using a binary cross-entropy loss in order to correctly predict whether the input refers to a real or a fake sample. The generator, on the other hand, is trained to generate samples as similar as possible to the real ones in order to fool the discriminator prediction. However, what would happen if we force the fake samples to fool the OVANet actual state, that is, be aligned with some category in $L_s$ by the classifier $C$ and recognized as ``known'' by the binary classifiers in $O$? We achieve this by adding two new constraints in the loss function of the generator: (\textit{i}) an entropy factor $\mathcal{L}_{ent}^{gen}$ (Equation~\ref{eq:entropy-gen}) based on the classifier $C$; and (\textit{ii}) an agreement factor $\mathcal{L}_{agree}^{gen}$ (Equation~\ref{eq:agreement-gen}) based on the binary classifiers of $O$.
 
\begin{equation} \label{eq:entropy-gen}
    \mathcal{L}_{ent}^{gen} = -\frac{1}{|L_s|} \sum_{k=1}^{|L_s|} p_c(y^k|\mathbf{\overline{x}}_i)\log(p_c(y^k|\mathbf{\overline{x}}_i))
\end{equation}

\begin{equation} \label{eq:agreement-gen}
    \mathcal{L}_{agree}^{gen} = -\frac{1}{|L_s|}\sum_{k=1}^{|L_s|}\log(p_o(\hat{y}^k|\mathbf{\overline{x}}_i))
\end{equation}
\end{enumerate}

\section{Experiments and Results} \label{sec:experiments}

In this section, we first provide details about the benchmark datasets. Then, we describe the experimental setup adopted to assess all three approaches. Finally, our final results are reported and compared with the baseline.

\subsection{Datasets}

We evaluate each of the proposed approaches described in Section~\ref{sec:negatives} on the Office-31~\cite{ECCV_2010_Saenko} and Office-Home~\cite{CVPR_2017_Venkateswara} datasets.
Office-31~\cite{ECCV_2010_Saenko} is composed of 4,652 organized into 31 classes of common objects from an office.
All images were collected from three different domains, named Amazon (A), 
Webcam (W), 
and DSLR (D). 
Office-Home~\cite{CVPR_2017_Venkateswara} consists of 15,500 images distributed into 65 categories and collected from several search engines and image directories.
It contains 4 distinct domains with strong domain shifts: Art (Ar), Clipart (Cl), Product (Pr), and Real World (Rw).

Following the same experimental protocol of Saito~and~Saenko~\cite{Saito_2021_ICCV}, both datasets were split into known ($L_s$) and unknown ($L_{unk}$) categories.
First, the labels of each dataset were sorted in alphabetical order. Then, the first 10 classes were selected for the known set and the remaining 21 classes for the unknown set in Office-31; whereas the first 20 classes were chosen for the known set and the last 45 classes for the unknown set in Office-Home.

\subsection{Evaluation Metrics}



OSDA methods are typically evaluated and compared by accuracy ($\textnormal{Acc}$), a performance measurement for classification problems that assess the proportion of examples correctly classified by a model.
However, Bucci~\etal~\cite{ECCV_2020_Bucci} have proved that accuracy is not a good measure to assess general OSDA methods, since it strongly depends on class balancing.
Indeed, we do not know beforehand what and how many unknown categories we are going to deal with.
Thus, they propose ``H-score'' ($\textnormal{Hsc}$), which is the harmonic mean between the accuracy for the known categories ($L_s$) and the accuracy for the unknown categories ($L_{unk}$) and is given by Equation~\ref{eq:hscore}.
\begin{equation} \label{eq:hscore}
    \textnormal{H-Score} = 2 \cdot \frac{\textnormal{Acc}^{L_s} \cdot \textnormal{Acc}^{L_{unk}}}{\textnormal{Acc}^{L_s} + \textnormal{Acc}^{L_{unk}}}
\end{equation}


\subsection{Experimental Protocol}

To ensure a fair comparison, we followed the same experimental protocol of OVANet~\cite{Saito_2021_ICCV}.
In each experiment, two domains were taken from a given dataset, one domain was chosen as source and the other as target, and evaluated using a full protocol setting~\cite{ICIAP_2017_Carlucci}, where the known set for the source domain (with labels) and the known and unknown sets for the target domain (without labels) were used for training and all the target samples from both its known and unknown sets were used for testing.
This procedure was repeated 3 times in order to ensure statistically solid results and the mean and standard deviation of the evaluation metrics were reported in the results.
All possible combinations of source and target domains were evaluated in our experiments.


\subsection{Implementation Details}

All three approaches presented in Section~\ref{sec:negatives} were implemented upon the original code of OVANet.
To do so, the code was first updated to use a newer version of PyTorch (1.13.0) and TorchVision (0.14.0) supporting our GPUs (\ie, compatible with CUDA 11.7) and then modified with as minimum changes as possible in order to implement all the aforesaid methods\iffinal\footnote{Our code is available at \url{https://github.com/jurandy-almeida/OVANet}}\fi.
This ensures we follow the same experimental procedure and set of hyper-parameters used for the original work, ensuring a fair comparison between the new results and the results already reported by its authors. 
For the \textit{augmentation} approach, we apply random affine transformations (as proposed in~\cite{ICLR_2018_French}) and Gaussian blur ($\sigma = 0.1$).
For the \textit{generation} approach, we implement the DCGAN from Chen~\etal~\cite{TPAMI_2021_Chen}. 
The hyperparameters used for training OVANet and all three approaches are presented in Table~\ref{tab:hyperparameters}.

\begin{table}[!htb]
    \centering
    \caption{Adopted Hyperparameters}
    \label{tab:hyperparameters}
    \begin{tabular}{r|c|c} \toprule
        Hyperparameters & Office-31 & Office-Home \\ \midrule 
        Feature Extractor               & ResNet50   & ResNet50   \\
        Source Known Classes            & $L_s = 10$ & $L_s = 25$ \\
        Target Unknown Classes          & $L_{unk} = 21$ & $L_{unk} = 40$ \\
        Optimizer                       & SGD        & SGD        \\
        batch-size                      & $36$       & $36$       \\
        Learning rate (newly instantiated)& $0.01$   & $0.01$     \\
        Learning rate (pre-trained)     & $0.001$    & $0.001$    \\
        Negative weight $\lambda$       & $0.1$      & $0.1$      \\ \midrule
        Iterations (pre-train)          & \multicolumn{2}{c}{$5000$}\\
        Additional Iterations (Original)           & \multicolumn{2}{c}{$0$}   \\
        Additional Iterations (Augmentation)       & \multicolumn{2}{c}{$0$}   \\
        Additional Iterations (Generation)         & \multicolumn{2}{c}{$1000$}\\
        Final Iterations (OVANet + approach)  & \multicolumn{2}{c}{$5000$}\\ \bottomrule
    \end{tabular}
\end{table}



\subsection{Ablation Study}

\begin{figure*}[!htb]
    \centering
    \subfloat[Unknown only]{
    \includegraphics[width=0.45\textwidth]{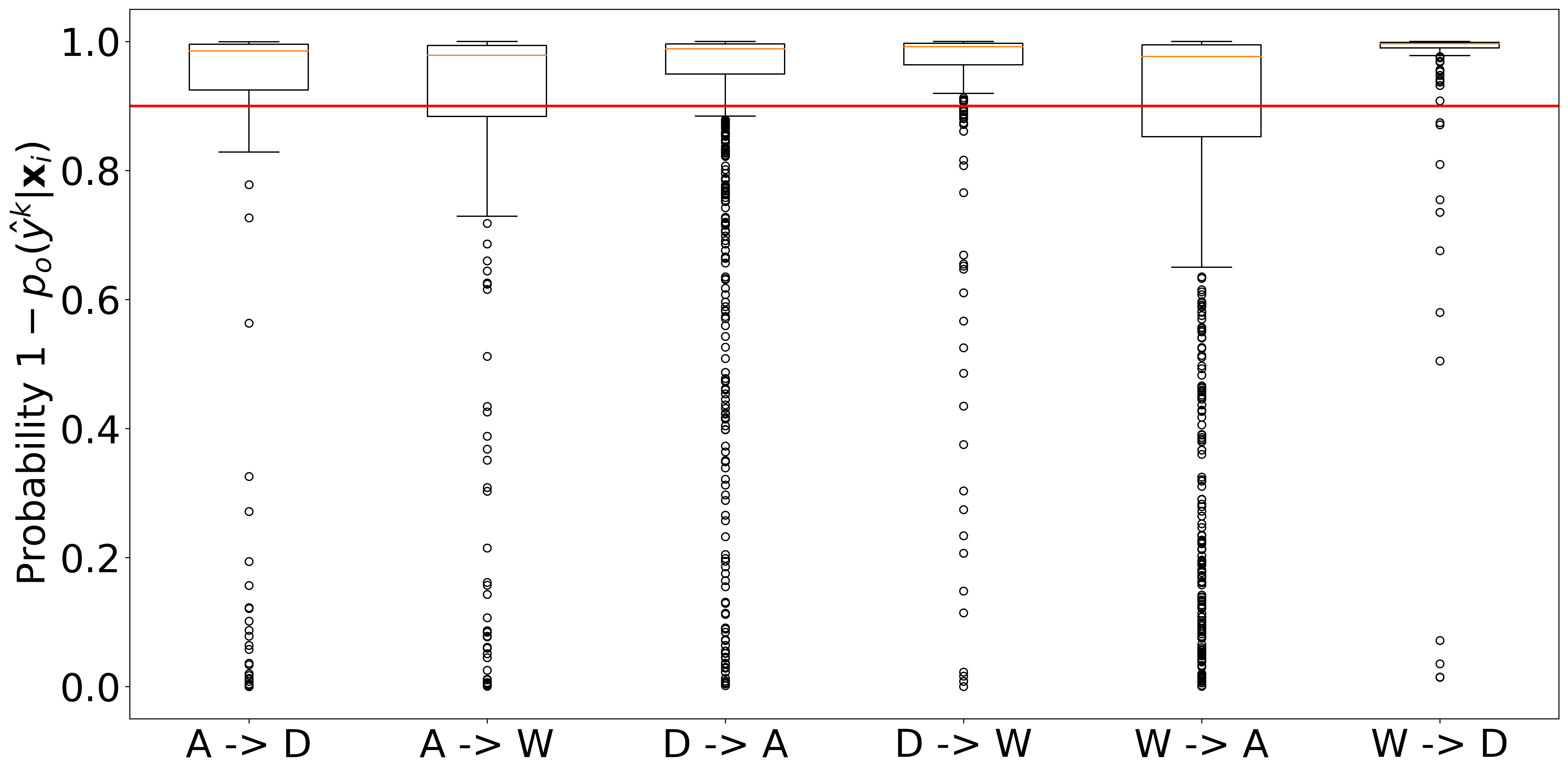}\label{fig:probs:unk}}
    \qquad
    \subfloat[Known only]{\includegraphics[width=0.45\textwidth]{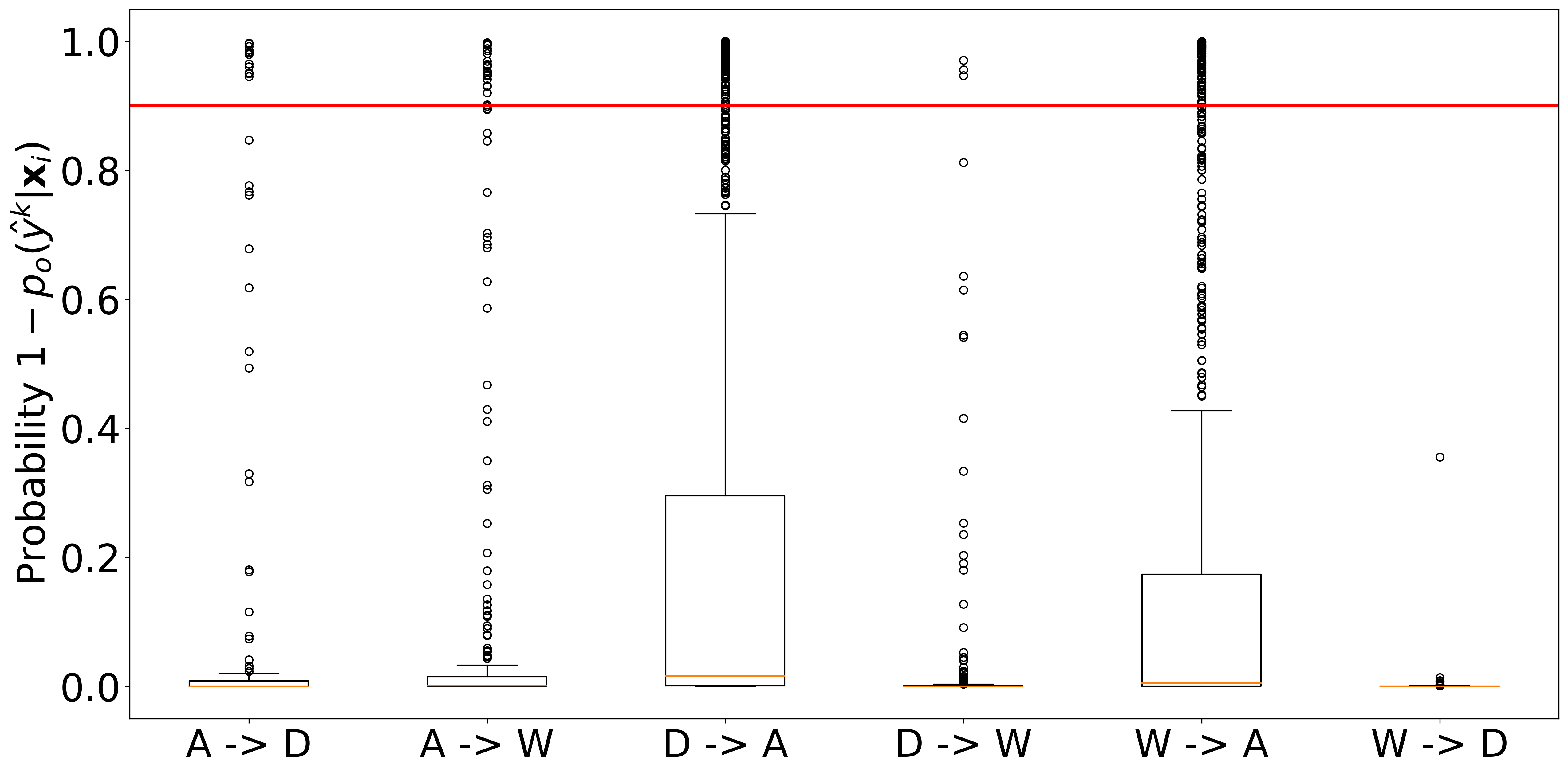}\label{fig:probs:knw}}
    \caption{Distribution of the obtained probabilities for (a) only unknown instances and (b) only known instances on the target training set of the Office-31 dataset. Each of the boxplots refers to the distribution of probabilities ($1-p_o(\hat{y}^k|\mathbf{x}_i)$) of a specific task. The red line shows the adopted threshold of 0.9.}
    \label{fig:probs}
\end{figure*}

We argue that it is important to draw high-confident negative instances from the target domain. 
To achieve this, we increase the threshold for rejecting unknown samples to a higher value close to 1.
Specifically, we set a threshold value of $0.9$. Figure~\ref{fig:probs:unk} shows that OVANet assigns $1-p_o(\hat{y}^k|\mathbf{x}_i) \geq 0.9$ for the majority of unknown samples of $\mathcal{D}_t$.
In contrast, Figure~\ref{fig:probs:knw} depicts that the probability $1-p_o(\hat{y}^k|\mathbf{x})$ for known samples is much lower than the adopted threshold.
Indeed, there may be outliers, but they are minimal compared to the vast majority of instances.
Therefore, 0.9 seems a suitable threshold choice.

\begin{figure*}[!htb]
    \centering

    \subfloat[Amazon]{\includegraphics[width=0.3\textwidth]{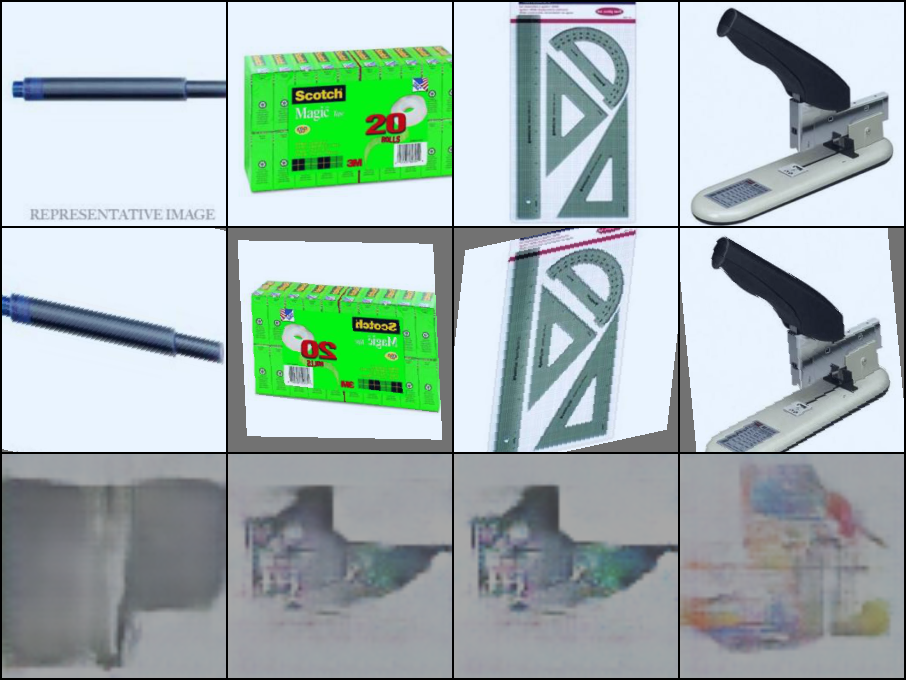}%
    \label{fig:negative-examples:amazon}}
    \hfill
    \subfloat[DSLR]{\includegraphics[width=0.3\textwidth]{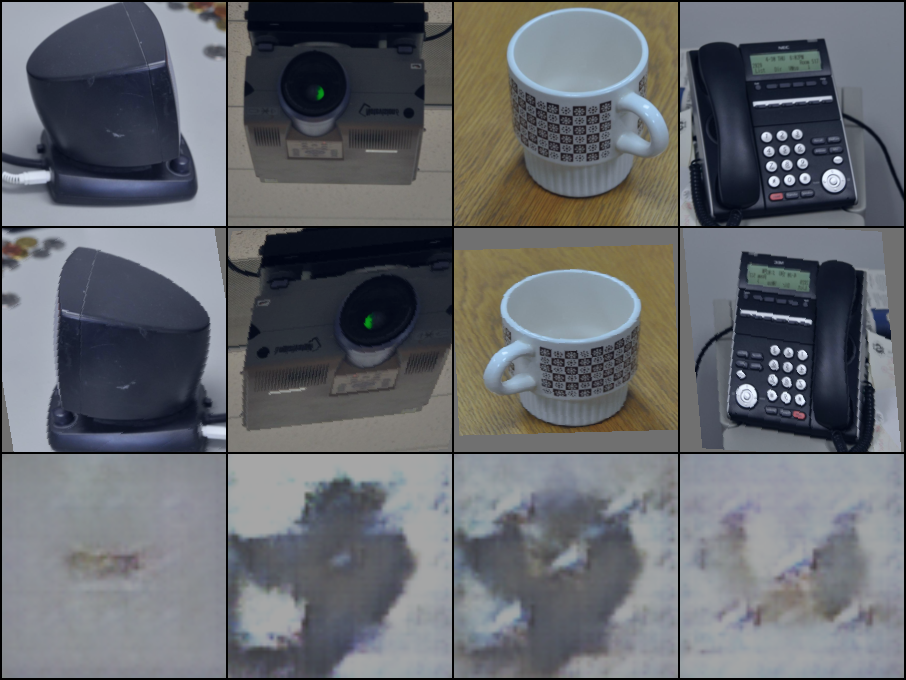}%
    \label{fig:negative-examples:dslr}}
    \hfill
    \subfloat[Webcam]{\includegraphics[width=0.3\textwidth]{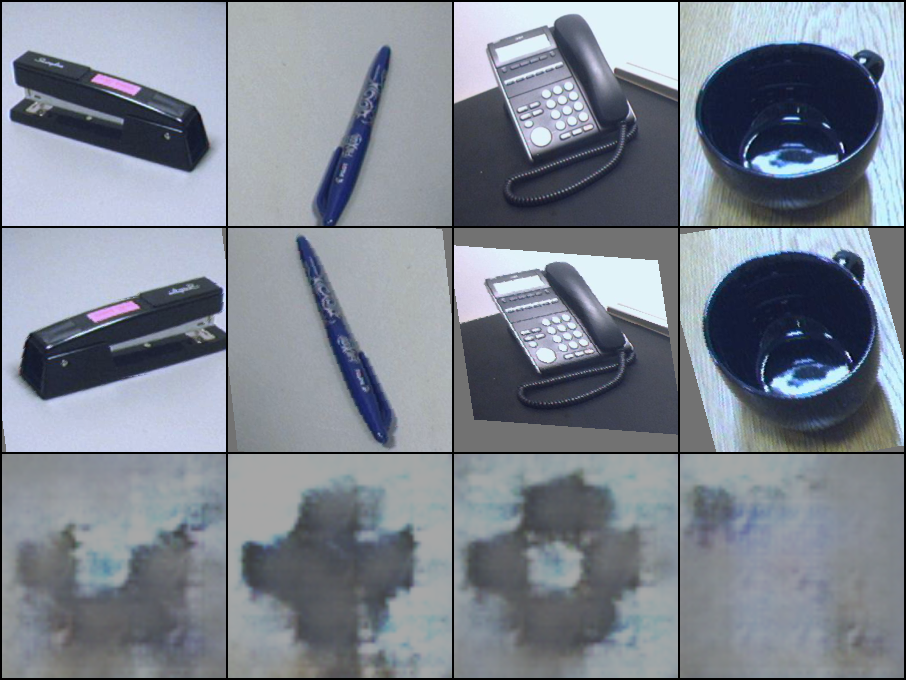}%
    \label{ffig:negative-examples:webcam}}
    
    \caption{Example of pristine, augmented, and generated negative samples. For each domain of Office-31, \ie, Amazon (a), DSLR (b), and Webcam (c), the first row contains examples of negative images, the second row depicts the augmented images, and, the last row shows the synthetically generated images.}
    \label{fig:negative-examples}
    


\end{figure*}

\subsection{Results}

In this section, we present and discuss the results obtained by the \textit{original}, \textit{augmentation}, and \textit{generation} approaches.
First of all, it is worth noting that, for the OSDA setting, Saito~and~Saenko~\cite{Saito_2021_ICCV} reported complete results only for Office-31 (see Table~B of~\cite{Saito_2021_ICCV}). 
For Office-Home, they present the result only for the task Rw\ra Ar (see Table~4 of~\cite{Saito_2021_ICCV}).
Thus, our very first step in order to have a solid baseline for comparison was to reproduce the OVANet results for Office-31 and Office-Home.

Table~\ref{tab:office31}~and~\ref{tab:officehome} present the results for the baseline (\ie, Reproducibility) and all three approaches on the Office-31 and Office-Home datasets, respectively.
The best result for each task is highlighted in bold.
Although the results are very task dependent, the use of unknown exploitation seems promising. 
Note that the baseline got the best result on only a few tasks.

\paragraph{Original} 
The first row of Figure~\ref{fig:negative-examples} illustrates examples of unknown samples from Office-31.
In general, it was possible to observe some small improvements compared to the baseline (\ie, Reproducibility). 
For instance, the task A\ra D got a 1\% increase in both Accuracy and H-score on Office-31.
On Office-Home, we observed up to 1.6\% of absolute gains in Accuracy for the task Rw\ra Cl.

\paragraph{Augmentation} 
Randomly perturbed versions (second row) for pristine unknown samples (first row) from Office-31 are depicted in Figure~\ref{fig:negative-examples}. 
On Office-31, the augmentation approach achieved the best result for the task W\ra D, with absolute gains of 1.3\% for both  Accuracy and H-Score.
On Office-Home, we could see about a 1\% increase over baseline (\ie, Reproducibility) for the task  Cl\ra Pr.

\paragraph{Generation} 
In the last row of Figure~\ref{fig:negative-examples}, there are examples of synthetic samples generated by DCGAN for each domain of Office-31 when taken as target.
Even though they are not very semantic for us, these examples were surgically made to be adversarial regarding OVANet and the discriminator. 
On Office-31, improvements of up to 1\% in both Accuracy and H-score over the baseline (\ie, Reproducibility) were observed. 
On Office-Home, there is more variance among the results, but nonetheless, absolute gains of 5.8\% in Accuracy and 4.7\% in H-score were achieved for the task Rw\ra Pr. 


\begin{table*}[!htb]
    \caption{Classification (\%) results for the Office-31 dataset.}
    \label{tab:office31}
    \centering
    \begin{tabular}{l|cc|cc|cc|cc} 
        \toprule
        \multirow{3}{*}{Tasks} & \multicolumn{8}{c}{Approaches} \\
        \cmidrule{2-9}
        & \multicolumn{2}{c|}{Reproducibility} & \multicolumn{2}{c|}{Original} & \multicolumn{2}{c|}{Augmentation} & \multicolumn{2}{c}{Generation} \\
        & Acc & Hsc & Acc & Hsc & Acc & Hsc & Acc & Hsc \\
        \midrule
        A\ra D & 86.8 $\pm$ 0.2 & 88.4 $\pm$ 0.2 
               & \textbf{87.8 $\pm$ 1.1} & \textbf{89.3 $\pm$ 1.0} 
               & 86.3 $\pm$ 1.0	& 87.9 $\pm$ 1.0 
               & 87.0 $\pm$ 1.2 & 88.6 $\pm$ 1.1 \\
        A\ra W & 86.9 $\pm$ 1.5 & 87.4 $\pm$ 1.4 
               & 87.2	$\pm$ 0.8 &	87.8 $\pm$ 0.8 
               & 86.2 $\pm$ 0.4 & 86.9 $\pm$ 0.5 
               & \textbf{87.8 $\pm$ 0.5} & \textbf{88.4 $\pm$ 0.3} \\
        D\ra A & 87.2 $\pm$ 1.0 & 86.4 $\pm$ 1.3 
               & \textbf{87.6	$\pm$ 0.5} & \textbf{87.0 $\pm$ 0.5} 
               & 85.9 $\pm$ 1.9 & 85.2 $\pm$ 2.2 
               & 81.7 $\pm$ 2.8 & 78.9 $\pm$ 3.8 \\
        D\ra W & 97.1 $\pm$ 1.1 & 97.3 $\pm$ 1.1 
               & \textbf{97.2	$\pm$ 0.6} & \textbf{97.4 $\pm$ 0.5} 
               & 97.0 $\pm$ 0.4 & 97.1 $\pm$ 0.5 
               & \textbf{97.2 $\pm$ 0.4} & \textbf{97.4 $\pm$ 0.2} \\
        W\ra A & 87.8 $\pm$ 1.0 & 87.9 $\pm$ 0.9 
               & 86.9	$\pm$ 0.8 & 86.9 $\pm$ 0.8 
               & \textbf{88.2 $\pm$ 0.7} & \textbf{88.1 $\pm$ 0.6} 
               & 84.7 $\pm$ 2.8 &	83.1 $\pm$ 4.0 \\
        W\ra D & 97.7 $\pm$ 0.6 & 97.8 $\pm$ 0.7 
               & 98.8 $\pm$ 0.3 & 98.8 $\pm$ 0.3 
               & \textbf{99.0 $\pm$ 0.3} & \textbf{99.0 $\pm$ 0.3} 
               & 98.3 $\pm$ 0.2 & 98.5 $\pm$ 0.2 \\
         \bottomrule
    \end{tabular}
\end{table*}


\begin{table*}[!htb]
    \caption{Classification (\%) results for the Office-Home dataset.}
    \label{tab:officehome}
    \centering
    \begin{tabular}{l|cc|cc|cc|cc} 
        \toprule
        \multirow{3}{*}{Tasks} & \multicolumn{8}{c}{Approaches} \\
        \cmidrule{2-9}
        & \multicolumn{2}{c|}{Reproducibility} & \multicolumn{2}{c|}{Original} & \multicolumn{2}{c|}{Augmentation} & \multicolumn{2}{c}{Generation} \\
        & Acc & Hsc & Acc & Hsc & Acc & Hsc & Acc & Hsc \\
        \midrule
        Ar\ra Cl & 64.5 $\pm$ 0.2	& \textbf{57.7 $\pm$ 0.5}
               & \textbf{65.5 $\pm$ 0.2} & 57.3 $\pm$ 0.2
               & 65.1 $\pm$ 0.2 & 56.0 $\pm$ 0.2
               & 63.6 $\pm$ 0.2 & 51.1 $\pm$ 1.3 \\
        Ar\ra Pr & 65.8 $\pm$ 0.5 & \textbf{66.7 $\pm$ 0.5} 
               & 65.6 $\pm$ 0.3 & 66.3 $\pm$ 0.3
               & 66.0 $\pm$ 0.4 & 66.5 $\pm$ 0.3
               & \textbf{67.6 $\pm$ 1.2} & 64.3 $\pm$ 2.3 \\
        Ar\ra Rw & 69.8 $\pm$ 0.1 & 69.3 $\pm$ 0.1
               & 70.6 $\pm$ 0.6 & 70.0 $\pm$ 0.7 
               & 71.1 $\pm$ 0.4 & 70.7 $\pm$ 0.4 
               & \textbf{73.0 $\pm$ 0.8} & \textbf{72.8 $\pm$ 0.8} \\
        Cl\ra Ar & 68.2 $\pm$ 0.6 & \textbf{60.9 $\pm$ 0.9} 
               & 68.1 $\pm$ 0.7 & 60.7 $\pm$ 0.8 
               & 67.6 $\pm$ 0.4 & 60.2 $\pm$ 0.4
               & \textbf{68.8 $\pm$ 0.1} & 59.7 $\pm$ 0.4 \\
        Cl\ra Pr & 66.7 $\pm$ 0.4 & 65.1 $\pm$ 0.3 
               & 67.4 $\pm$ 0.2 & 65.5 $\pm$ 0.3  
               & \textbf{67.5 $\pm$ 0.6} & \textbf{65.7 $\pm$ 0.5}  
               & 67.1 $\pm$ 0.3 & 64.7 $\pm$ 0.1 \\
        Cl\ra Rw & 67.8 $\pm$ 0.4 & 67.5 $\pm$ 0.3 
               & 68.0 $\pm$ 0.7 & \textbf{67.6 $\pm$ 0.5} 
               & 68.2 $\pm$ 0.6 & \textbf{67.6 $\pm$ 0.6}
               & \textbf{68.6 $\pm$ 0.3} & 67.5 $\pm$ 0.4 \\
        Pr\ra Ar & 66.5 $\pm$ 0.7 & \textbf{59.3 $\pm$ 0.8} 
               & 67.6 $\pm$ 0.9 & 59.1 $\pm$ 1.0  
               & 67.0 $\pm$ 0.3 & 58.5 $\pm$ 0.6  
               & \textbf{67.9 $\pm$ 0.4} & 57.0 $\pm$ 0.6 \\
        Pr\ra Cl & 63.2 $\pm$ 0.2 & \textbf{52.2 $\pm$ 0.7}  
               & \textbf{64.1 $\pm$ 0.2} & 51.5 $\pm$ 0.4  
               & 63.3 $\pm$ 0.2 & 50.2 $\pm$ 0.5
               & 61.7 $\pm$ 0.3 & 44.0 $\pm$ 1.8 \\
        Pr\ra Rw & 69.3 $\pm$ 1.1 & 68.9 $\pm$ 1,1 
               & 69.3 $\pm$ 0.3 & 68.9 $\pm$ 0.2  
               & 70.4 $\pm$ 0.3 & 69.9 $\pm$ 0.2 
               & \textbf{71.8 $\pm$ 0.8} & \textbf{70.9 $\pm$ 0.7} \\
        Rw\ra Ar & 68.6 $\pm$ 0.4 & 67.6 $\pm$ 0.3  
               & \textbf{69.9 $\pm$ 0.8} & \textbf{68.3 $\pm$ 0.8}
               & 69.0 $\pm$ 0.2 & 67.5 $\pm$ 0.5 
               & 68.8 $\pm$ 0.8 & 66.7 $\pm$ 0.5 \\
        Rw\ra Cl & 62.5 $\pm$ 0.7 & \textbf{58.6 $\pm$ 0.6} 
               & \textbf{64.1 $\pm$ 0.2} & 58.3 $\pm$ 0.4 
               & 63.5 $\pm$ 0.7 & 57.2 $\pm$ 0,1 
               & 61.5 $\pm$ 0.7 & 51.5 $\pm$ 1.1  \\
        Rw\ra Pr & 66.1 $\pm$ 0.2 & 66.4 $\pm$ 0.3  
               & 67.0 $\pm$ 0.5 & 67.5 $\pm$ 0.6
               & 67.1 $\pm$ 0.8 & 67.7 $\pm$ 1.0  
               & \textbf{71.9 $\pm$ 0.6} & \textbf{71.1 $\pm$ 0.1}  \\
         \bottomrule
    \end{tabular}
\end{table*}

\section{Conclusion} \label{sec:conclusion}

Deep models over the last few years have shown outstanding results in many challenging tasks.
However, when they are deployed to uncontrolled environments, such as the real world, they often struggle when faced with unsupervised datasets, possibly achieving under-expected performance because of domain shift and category shift problems.
Particularly, the research field of OSDA has been proposing methods to tackle those problems by recognizing unknown samples that may appear during inference.
We present three new approaches to improve OSDA methods by using unknown samples to tighten the classification boundaries of the model.
The use of our approaches may increase performance in most OSDA tasks, reaching up to 1.3\% of absolute gains for both Accuracy and H-Score on Office-31 and 5.8\% for Accuracy and 4.7\% for H-Score on Office-Home.
As future work, we envisage the investigation of newer GAN models to generate high-resolution negative images and the assessment of the model in different datasets, such as DomainNet~\cite{ICCV_2019_Peng} and CerraData~\cite{SIBGRAPI_2022_Miranda}.

\iffinal
\section*{Acknowledgments}
This research was supported by São Paulo Research Foundation - FAPESP (2020/08770-3, 2021/13348-1, 2023/03328-9), FAPESP-Microsoft Research Institute (2017/25908-6), Brazilian National Council for Scientific and Technological Development - CNPq (314868/2020-8), and LNCC via resources of the SDumont supercomputer of the IDeepS project.
\fi





\end{document}